\documentclass[twocolumn,showpacs,%
  nofootinbib,aps,superscriptaddress,%
  eqsecnum,prd,notitlepage,showkeys,10pt]{revtex4-1}

\usepackage{amssymb}
\usepackage{amsmath}
\usepackage{graphicx}
\usepackage{dcolumn}
\usepackage{hyperref}
\usepackage{adjustbox}
\usepackage{subcaption}
\usepackage{float}

\usepackage{titlesec}
\titlespacing*{\section}{\baselineskip}{1.4\baselineskip}{\baselineskip}
\titlespacing*{\subsection}{\baselineskip}{\baselineskip}{0.7\baselineskip}

\begin{document}

\renewcommand{\topfraction}{.85}
\renewcommand{\bottomfraction}{.7}
\renewcommand{\textfraction}{.15}
\renewcommand{\floatpagefraction}{.66}
\renewcommand{\dbltopfraction}{.66}
\renewcommand{\dblfloatpagefraction}{.66}
\setcounter{topnumber}{9}
\setcounter{bottomnumber}{9}
\setcounter{totalnumber}{20}
\setcounter{dbltopnumber}{9}

\title{Image Outpainting and Harmonization using Generative Adversarial Networks}
\author{Basile Van Hoorick \\[0.05cm] \textit{Columbia University} \\[0.05cm] \texttt{basile.vanhoorick@columbia.edu}}

\begin{abstract}
\textbf{Abstract} \hspace{0.3cm}
Although the inherently ambiguous task of predicting what resides beyond all four edges of an image has rarely been explored before, we demonstrate that GANs hold powerful potential in producing reasonable extrapolations. Two outpainting methods are proposed that aim to instigate this line of research: the first approach uses a context encoder inspired by common inpainting architectures and paradigms, while the second approach adds an extra post-processing step using a single-image generative model. This way, the hallucinated details are integrated with the style of the original image, in an attempt to further boost the quality of the result and possibly allow for arbitrary output resolutions to be supported.
\end{abstract}

\maketitle

\begin{figure}[tp]
\centering
  \begin{adjustbox}{center}
  \includegraphics[width=\columnwidth]{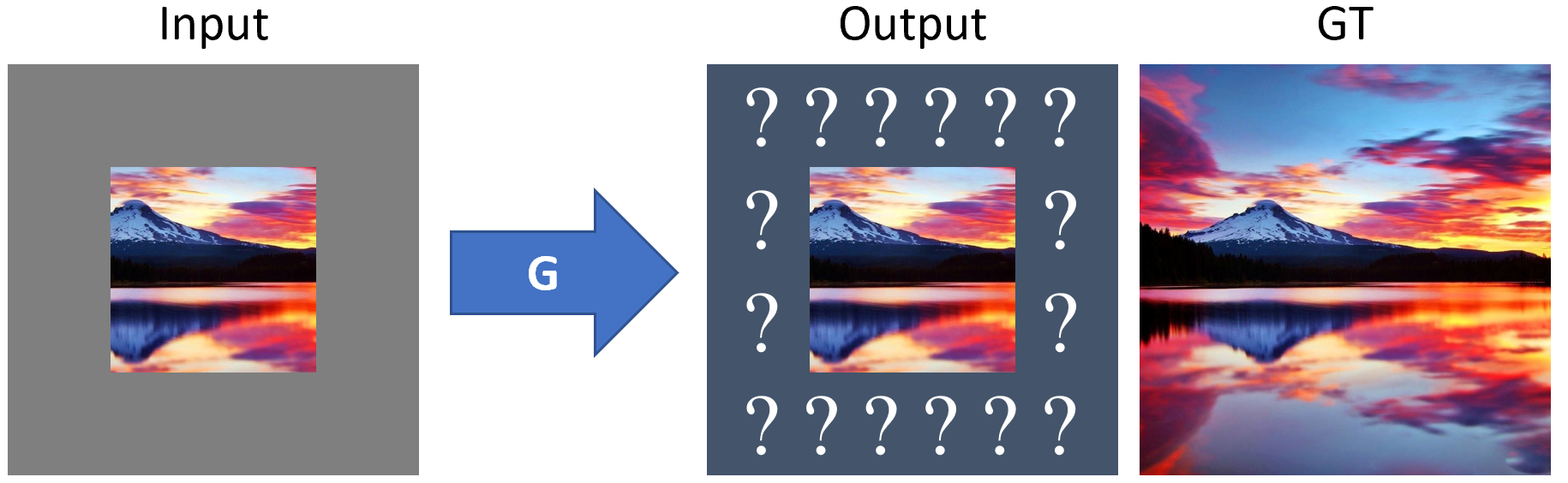}
  \end{adjustbox}
  \caption{Image outpainting idea.}
  \label{fig:GenIdea}
\end{figure}

\section{INTRODUCTION}

\begin{figure}[tp]
\centering
  \begin{adjustbox}{center}
  \includegraphics[width=\columnwidth]{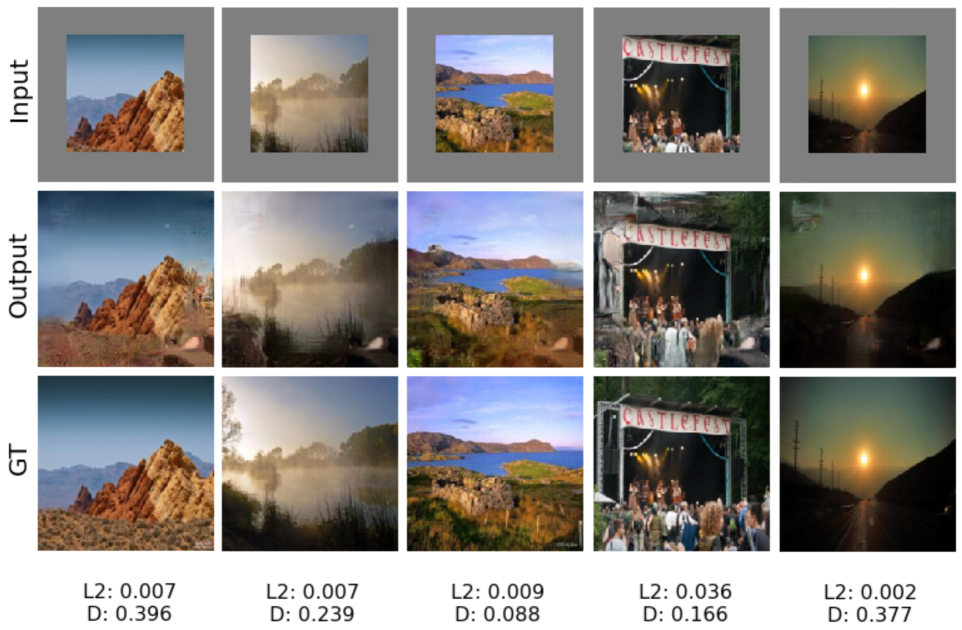}
  \end{adjustbox}
  \caption{Demonstration of the outpainting task, as achieved using an adversarially trained generator.}
  \label{fig:B3_samples}
\end{figure}

\hspace{\parindent} When presented with an incomplete image, humans are excellent at filling in the blanks and producing a realistic explanation for what could be missing. Image inpainting is a well-studied problem that replicates this behaviour, often tasking deep neural networks with trying to understanding the semantic content of natural images in order to recover the missing regions of a photo. However, the spatially inverted variant of this problem is even more challenging and, with a small play on words, can be denoted \textit{outpainting}. The problem statement is shown in Figure \ref{fig:GenIdea}; essentially, the task is to extrapolate the image content rather than to interpolate within an image. More formally, we must design a generator $G$ that converts an image $x$ with dimensions $n*n$ into a larger image $G(x)$ with dimensions $m*m$, such that the center part of $G(x)$ looks the same as $x$, while the complete outpainted image $G(x)$ should be a plausible hypothesis of what could encompass the original image. In particular, the generated parts should look realistic, despite the fact that it is often impossible to know precisely what the scene contains outside of the pictured boundaries. Figure \ref{fig:B3_samples} gives a practical illustration of what is accomplished in this work. Possible applications
of outpainting include the opportunity to experience and explore multimedia more immersively, enabling technologies similar to Ambilight to produce much richer and/or realistic surroundings, or merely as a form of computer-generated art that can be appreciated aesthetically.

\subsection{Related Work}

Deep neural networks have recently shown great performance in image completion. This section discusses previous work related to the proposed methods for outpainting.

\textbf{Self-supervised learning} \hspace{0.3cm}
The automatic generation of a supervision signal, by systematically omitting certain parts of the input data, grants the ability to capitalize on the vast amounts of unlabeled images and videos that are available on the Web. For example, both colorization \cite{Larsson2017} and predicting the relative spatial position of two patches \cite{Doersch2015, Noroozi2016} have been proven to be successful pretext tasks to representation learning. These methods do not require any labellings of the dataset, since the inputs (gray-scale images resp. patches) as well as their corresponding ground truths (color images resp. locations) can be extracted procedurally.

\textbf{Generative Adversarial Networks} \hspace{0.3cm}
Image or video generation using convolutional neural networks with an adversarial loss \cite{Goodfellow2014} has attracted significant research interest. The idea is to let a \textit{generator} network $G$ create samples according to some distribution, and subsequently have an auxiliary network called the \textit{discriminator} $D$ try to distinguish whether a given sample is valid or was actually generated by $G$. These two components are trained in a quick alternation, where $G$ trying to fool $D$ forces $D$ to become better at telling real from fake, which in turn motivates $G$ to synthesize increasingly convincing outputs \cite{Iizuka2017}. From the perspective of game theory, GANs can be viewed as a zero-sum two-player game; as such, they are comparatively hard to train. In fact, it turns out that the conditions for their convergence is still an open research problem \cite{Mescheder2018}.

\textbf{Inpainting and Context Encoders} \hspace{0.3cm}
Recovering holes within images has important applications for the restoration of damaged media, and can be achieved by training GANs in a self-supervised way. Specifically, a generator can learn to 'fill in the blanks' by using an encoder-decoder network that encodes the surrounding context in order to understand the image content, and subsequently produces a plausible hypothesis for the missing square \cite{Pathak2016}. More recent improvements include attempts to complete images of arbitrary resolutions by filling in regions of any shape, and employing two types of discriminators (\textit{global} and \textit{local}) in order to enforce both overall consistency as well as realistic details \cite{Iizuka2017}.

\begin{figure}[tp]
\centering
  \begin{adjustbox}{center}
  \includegraphics[width=\columnwidth]{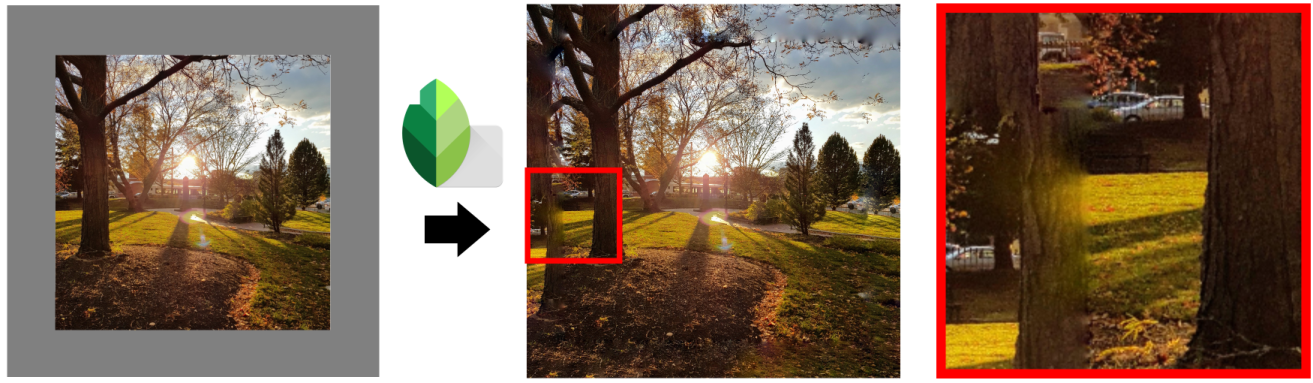}
  \end{adjustbox}
  \caption{Demonstrating Snapseed's \cite{snapseed} unrealistic results upon running the 'Expand' feature on a natural photograph.}
  \label{fig:snapseed_fail}
\end{figure}

\textbf{Outpainting} \hspace{0.3cm}
The problem opposite to inpainting comprises predicting which pixels reside beyond the borders of a fully intact photo, and has to our knowledge been explored only few times before by the academic research community \cite{Zhang2013, Sabini2018, Wang2019}. One approach geometrically extrapolates the field of view of an image using another panoramic reference image of the same scene category using old-school computer vision techniques \cite{Zhang2013}, while a more recent relevant paper uses a GAN to perform horizontal outpainting \cite{Sabini2018}. Results look promising, although there is room for visual improvement in terms of making the hallucinations omnidirectional and increasing the output resolution. \cite{Wang2019} achieves impressive outcomes but tends to focus on limited domain datasets (such as faces only), and notes that generative models experience difficulties trying to fit datasets as diverse as Places2. Lastly, Google's Snapseed application has a proprietary 'Expand' functionality that seems to select patches from the image and copy them to the edges \cite{snapseed}, but a limited number of experiments suggest that this tool fails to capture the local structure of most scenes from region to region; see Figure \ref{fig:snapseed_fail}.

\textbf{Single-image generative models} \hspace{0.3cm}
The recently introduced SinGAN framework can train an unconditional generative model on a single natural photo, capturing and reproducing image statistics across various scales of the image \cite{Shaham2019}. It allows for the creation of random samples with new object configurations by starting from a low-resolution sample at the coarsest scale, and then progressively upsampling and refining the result through the pyramid of generators. By carefully modifying the initial sample, SinGAN can be used to perform several image manipulation tasks such as editing photos, splicing foreign objects into the scene and subsequently harmonizing their style with the environment, or even superresolution.

\section{METHODS}

\subsection{Datasets}

\hspace{\parindent} During training, we decide to crop images first before feeding them into the generator. Consequently, learning can be done in a self-supervised way thanks to the fact that we are now able to enforce the output to approximate the original, uncropped image. Any sufficiently large dataset of unlabelled, natural photos will therefore suffice. We continue with the MIT CSAIL Places365-Standard dataset, which contains millions of images with landscapes, buildings and other everyday scenarios intended for scene recognition \cite{Zhou2018}. A second set of experiments was performed with a dataset consisting of images scraped from WikiArt \cite{wikiart}. This website holds a database of visual artwork belonging to many different categories.

The photos will first be resized to 192x192 as a preprocessing step, and the generator will then be tasked with expanding a crop of 128x128 back into a 192x192 image. Note that adding 32-pixels at every boundary might not seem significant, but we consider this configuration to be quite ambitious already: the total output size is 2.25 times that of the input, meaning that over half of all pixels will be hallucinated. In practice, the generator $G$ maps a partially masked 192x192 color image to an outpainted variant of the same dimensions, with the masked part replaced by the model's predictions.

\begin{figure*}[tp]
  \centering
  \begin{adjustbox}{center}
  \includegraphics[width=\textwidth]{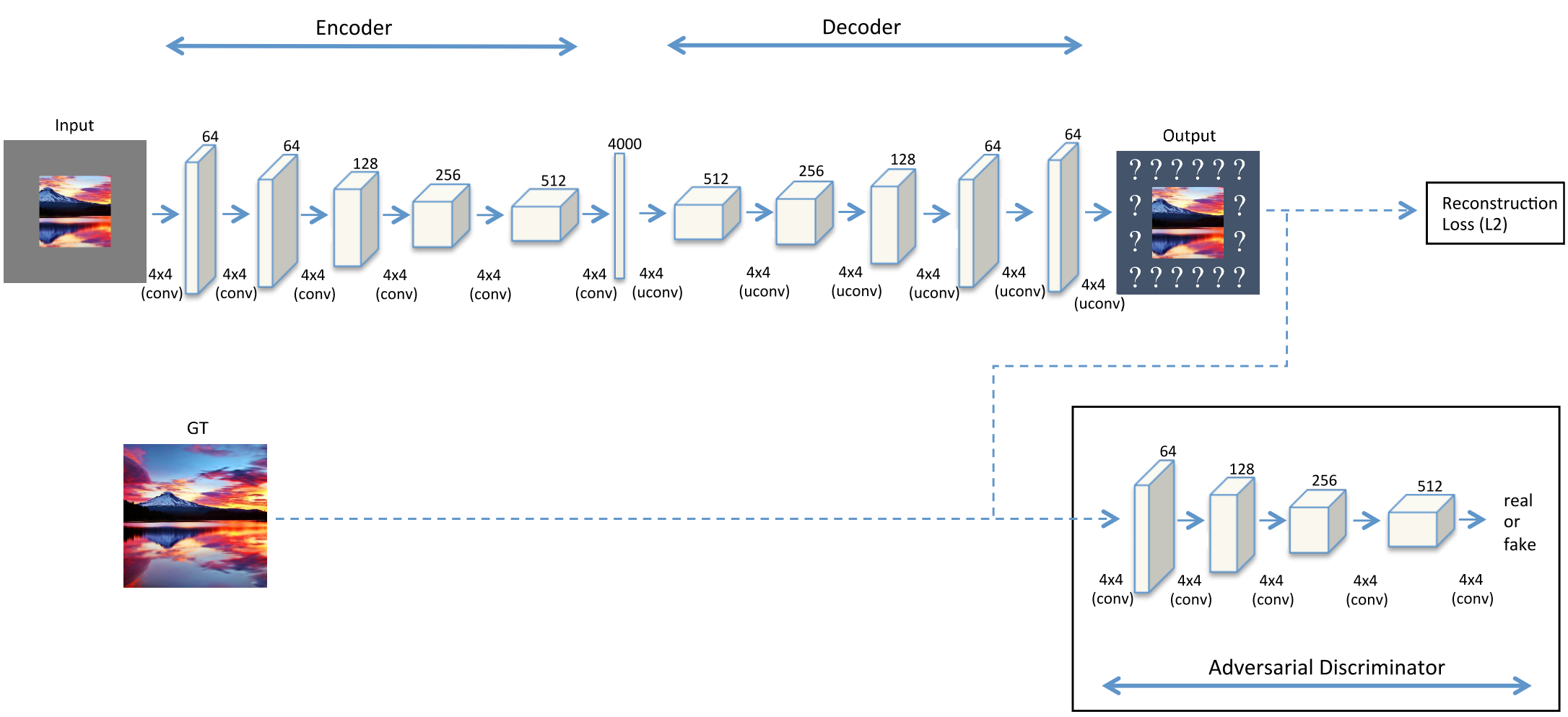}
  \end{adjustbox}
  \caption{Context encoder trained with joint reconstruction and adversarial loss for semantic outpainting, based on \cite{Pathak2016}. Note that as opposed to inpainting, the decoder is almost an exact mirroring of the encoder, since both input and output have the same dimensions.}
  \label{fig:MyArch}
\end{figure*}

\begin{table}[tp]
\centering
\footnotesize
\begin{tabular}{l c c}
  \hline
  Type & Kernel size & $(H,W,C)$ \\ \hline\hline
  Conv + Leaky ReLU & $4 * 4$ & $(96, 96, 64)$ \\
  Conv + BatchNorm + Leaky ReLU & $4 * 4$ & $(48, 48, 64)$ \\
  Conv + BatchNorm + Leaky ReLU & $4 * 4$ & $(24, 24, 128)$ \\
  Conv + BatchNorm + Leaky ReLU & $4 * 4$ & $(12, 12, 256)$ \\
  Conv + BatchNorm + Leaky ReLU & $4 * 4$ & $(6, 6, 512)$ \\
  Conv & $4 * 4$ & $(3, 3, 4000)$ \\
  Up-conv + BatchNorm + ReLU & $4 * 4$ & $(6, 6, 512)$ \\
  Up-conv + BatchNorm + ReLU & $4 * 4$ & $(12, 12, 256)$ \\
  Up-conv + BatchNorm + ReLU & $4 * 4$ & $(24, 24, 128)$ \\
  Up-conv + BatchNorm + ReLU & $4 * 4$ & $(48, 48, 64)$ \\
  Up-conv + BatchNorm + ReLU & $4 * 4$ & $(96, 96, 64)$ \\
  Up-conv + BatchNorm + Tanh & $4 * 4$ & $(192, 192, 3)$ \\ \hline
\end{tabular}
\caption{Layers of the generator network $G$.}
\label{tab:layers_G}
\end{table}

\begin{table}[tp]
\centering
\footnotesize
\begin{tabular}{l c c}
  \hline
  Type & Kernel size & $(H,W,C)$ \\ \hline\hline
  Conv + Leaky ReLU & $3 * 3$ & $(96, 96, 64)$ \\
  Conv + InstanceNorm + Leaky ReLU & $3 * 3$ & $(48, 48, 128)$ \\
  Conv + InstanceNorm + Leaky ReLU & $3 * 3$ & $(24, 24, 256)$ \\
  Conv + InstanceNorm + Leaky ReLU & $3 * 3$ & $(24, 24, 512)$ \\
  Conv & $3 * 3$ & $(24, 24, 1)$ \\ \hline
\end{tabular}
\caption{Layers of the discriminator network $D$.}
\label{tab:layers_D}
\end{table}

\subsection{Architecture}

\hspace{\parindent} Many architectural aspects and ideas can be naturally adopted from inpainting. The context encoder part of the generator network $G$ repeatedly downsamples the masked input through six convolutional layers, in order to efficiently capture the image content and object semantics within some embedding space. Next, the decoder consists of a special kind of layers called \textit{up-convolutional} or \textit{deconvolutional}, which can be understood as having a fractional stride in order to 'undo' the downsampling performed by the encoder \cite{Pathak2016}.

The discriminator $D$ is another deep neural network that estimates the probability of the ground truth or the hallucinated image being real. In inpainting, $D$ typically sees the generated part only \cite{Pathak2016}, although in this project we decide to operate on the full outpainted image in order to discourage $G$ from introducing obvious perceptual discontinuities or other kinds of structural inconsistencies. The architecture we use for $D$ results in a $24*24$ grid of probabilities whose errors are averaged during the training process.

See Figure \ref{fig:MyArch} as well as Tables \ref{tab:layers_G} and \ref{tab:layers_D} for a detailed overview of the system architecture.

\subsection{Training}

\hspace{\parindent} Training is done for 200 epochs, with a fixed learning rate of $\alpha=0.0003$ and two Adam optimizers with $\beta_1=0.5, \beta_2=0.999$. The loss functions are as follows:
\begin{align}
  L_{rec} &= ||x-G(x)||_1 \\
  L_{adv} &= ||D(G(x))-1||_2^2 \\
  L_G &= \lambda_{rec}L_{rec} + \lambda_{adv}L_{adv} \\
  L_D &= ||D(x)-1||_2^2 + ||D(G(x))-0||_2^2
\end{align}
Using an $L_1$ reconstruction loss instead of $L_2$ helps produce less blurry images \cite{Huang2018}. The weight of the adversarial loss $\lambda_{adv}$ relative to the reconstruction loss $\lambda_{rec}=1-\lambda_{adv}$ turned out to be particularly tricky to adjust; this factor was initially set to $\lambda_{adv}=0.001$ as in various existing works \cite{Pathak2016, Li2018, Sabini2018}, although the GAN kept collapsing into a failure mode where the adversarial loss did not move away from $1$. This means that $G$ was unable to fool $D$, so $D$ was ahead of $G$ and could always tell real from fake successfully. A working remedy involved varying $\lambda_{adv}(n)$ throughout time as a function of the epoch $n$ as follows:
\begin{align}
  \lambda_{adv}(n) = \begin{cases}
    0.001, & \text{if }n\leq10 \\
    0.005, & \text{if }10<n\leq30 \\
    0.015, & \text{if }30<n\leq60 \\
    0.040, & \text{otherwise}
  \end{cases}
  \label{eq:loss_adv}
\end{align}
This will punish the generator more heavily for producing unrealistic outputs as time progresses, rather than just enforcing an accurate pixel-wise reconstruction.

\subsection{Harmonization}

\hspace{\parindent} Due to the visual fuzziness of our initial experiments, we came up with the idea of leveraging SinGAN's harmonization capabilities in an attempt to improve the fidelity of the hallucinated outputs, serving as a second possible approach to the outpainting problem. To this end, we first train a SinGAN model on the original high-quality image, then propagate it forward through the outpainting generator (which produces a low-resolution output), and then try to super-resolve this result by injecting it into one of the coarser scales of SinGAN. The hope is that the model will harmonize the outpainted parts with the style of the original image that it was trained on, while simultaneously synthesizing a finer-scale, higher-resolution variant by pushing it up through the hierarchy of multi-scale generators.\footnote{Note that our definition of the term \textit{harmonization} is slightly different from what is actually meant in the referred paper, but the underlying process is exactly what we intend to perform anyway.}

\begin{figure}[tp]
  \centering
  \begin{subfigure}[b]{\columnwidth}
      \centering
      \includegraphics[width=\columnwidth]{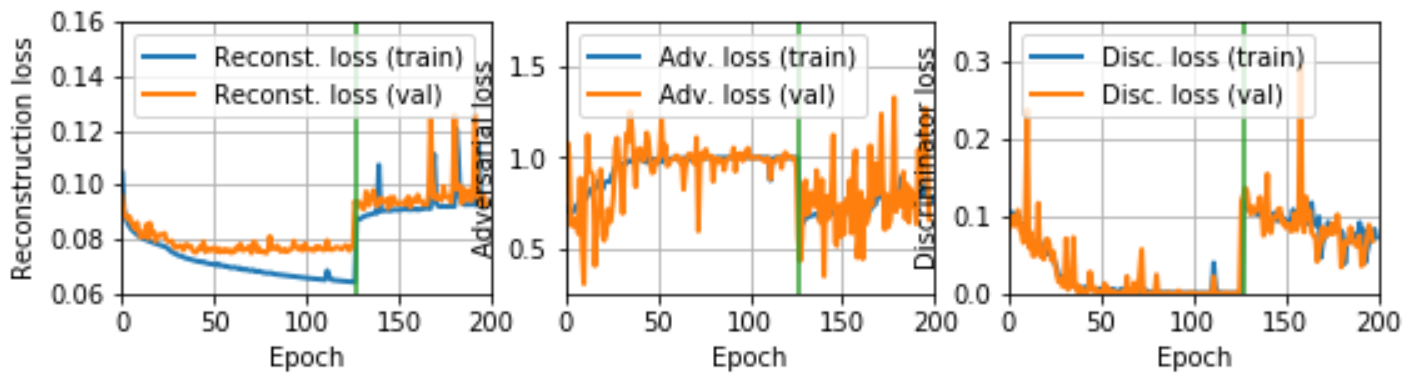}
      \caption{Places365 (green line marks restart).}
      \vspace{0.2cm}
  \end{subfigure}
  \begin{subfigure}[b]{\columnwidth}
      \centering
      \includegraphics[width=\columnwidth]{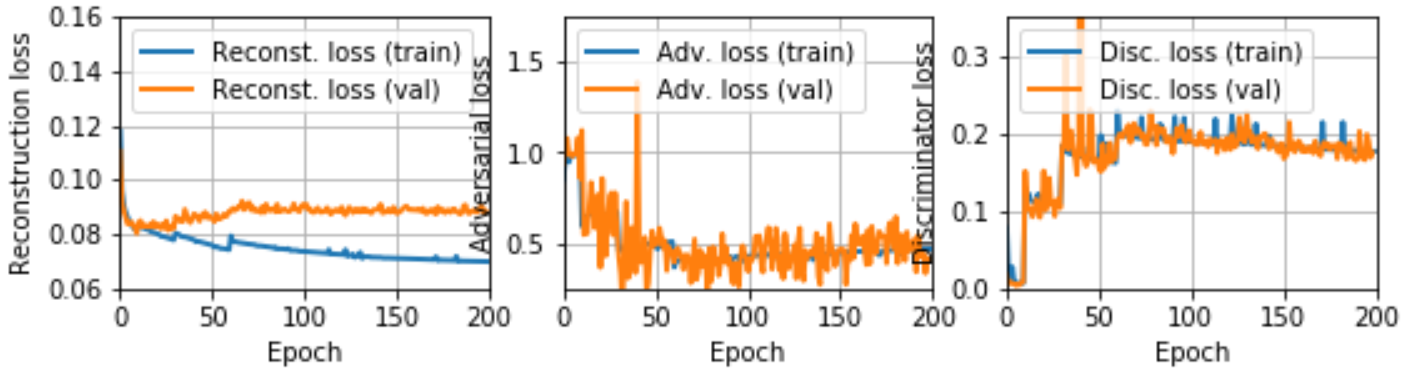}
      \caption{WikiArt.}
  \end{subfigure}
  \caption{Learning curves for the outpainting GAN, where the difficulty of training GANs is observed.}
  \label{fig:learn_curves}
\end{figure}

\section{RESULTS}

\subsection{Experiments}

\hspace{\parindent} Three models were trained and evaluated, differing in the following aspects:
\begin{itemize}
  \item 200k images of Places365, with $L_G = L_{rec}$, so only the $L_1$ reconstruction loss is considered.
  \item 200k images of Places365, with fully functional joint reconstruction and adversarial loss functions.
  \item 50k images of WikiArt, with fully functional joint loss functions.
\end{itemize}
Since training takes a long time, the second model's training process was momentarily interrupted and resumed in order to upgrade the adversarial loss weight to the better performing Equation \ref{eq:loss_adv}. Selected learning curves are plotted in Figure \ref{fig:learn_curves}. Note that the discriminator $D$ seems to have a harder time deciding for the WikiArt dataset than for Places365, both in terms of adversarial loss $L_{adv}$ and regular discriminator loss $L_D$. This suggests a priori that the artwork domain is inherently more diverse than natural scenery in a visual sense, making it harder to reliably classify the authenticity of samples.

\begin{table}[tp]
\centering
\begin{tabular}{l r r}
  \hline
  Model & Mean $MSE(G(x),y)$ & Mean $D(G(x))$ \\ \hline\hline
  Places365 rec & 0.0181 & 0.0791 \\
  Places365 rec + adv & 0.0230 & 0.1705 \\
  WikiArt rec + adv & 0.0227 & 0.2371 \\ \hline
\end{tabular}
\caption{Mean square error and realism (i.e. probability of legitimacy) statistics.}
\label{tab:quant}
\end{table}

\subsection{Evaluations}

\hspace{\parindent} In the final step of the outpainting pipeline, the input is pasted on top of and blended with the output using a simple 16-pixel wide gradient alpha mask. Compression artefacts arising from the autoencoder-like operation at the center of the image are thus resolved by simply hiding them. However, two quantitative performance metrics, the mean squared error and the discriminator's output values, are calculated from the original, unmodified result. See Table \ref{tab:quant} for an overview of these metrics for the three trained models as calculated on the respective test sets. Including the adversarial loss increases the reconstruction loss, and outpainted artwork fools the discriminator more often than when we stay within the domain of natural images.

\begin{figure*}[tp]
\centering
  \begin{adjustbox}{center}
  \includegraphics[width=\textwidth]{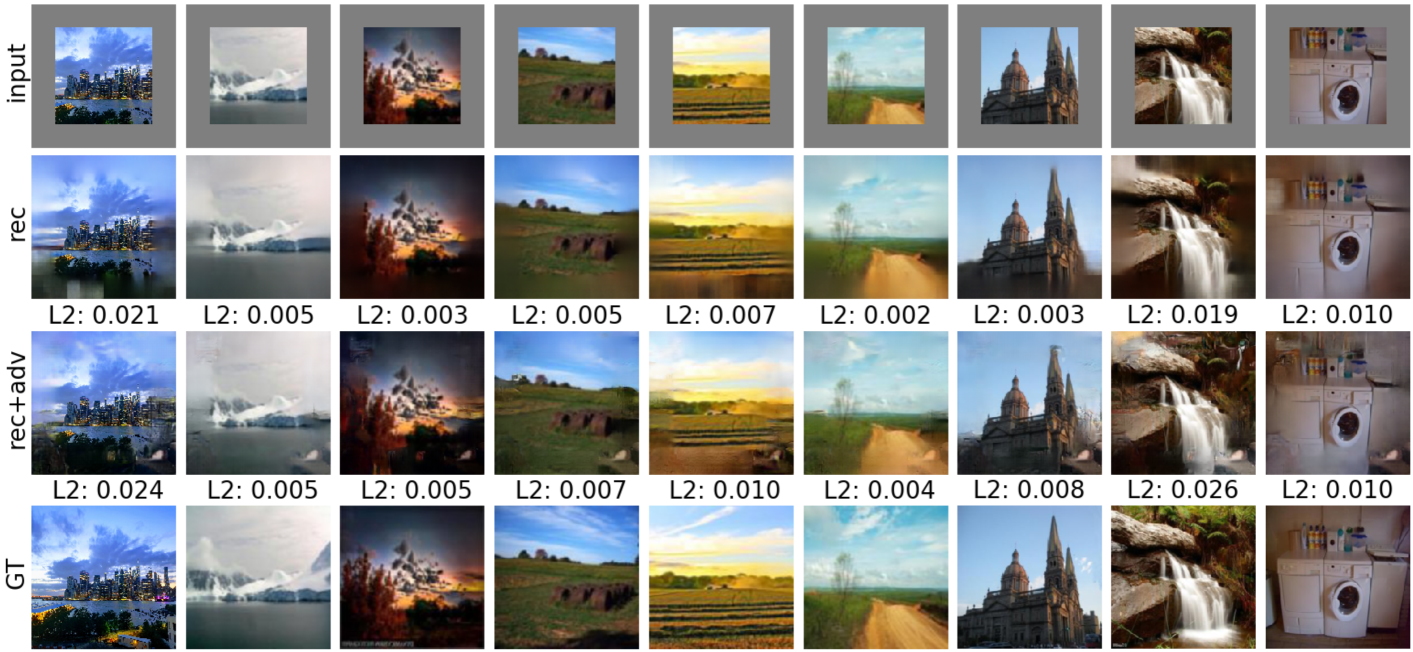}
  \end{adjustbox}
  \caption{Comparison between outpainting on Places365 without or with adversarial loss, demonstrating blurry results in the former case.}
  \label{fig:B2_B3_cmp}
\end{figure*}

\textbf{Non-adversarial versus adversarial models} \hspace{0.3cm}
See Figure \ref{fig:B2_B3_cmp} for samples of generators trained with $\lambda_{adv}=0$ and $\lambda_{adv}=0.040$ respectively. If only the $L_1$ reconstruction loss is used, then the hallucinated part looks fuzzy and unrealistic. Enabling the discriminator clearly has a sharpening effect on the output, although the results start to look excessively decorated at times. It seems that the trade-off between the reconstruction and adversarial loss functions should be more finely investigated.

\begin{figure*}[tp]
  \centering
  \begin{subfigure}[t]{0.5\textwidth}
      \centering
      \includegraphics[width=0.95\linewidth]{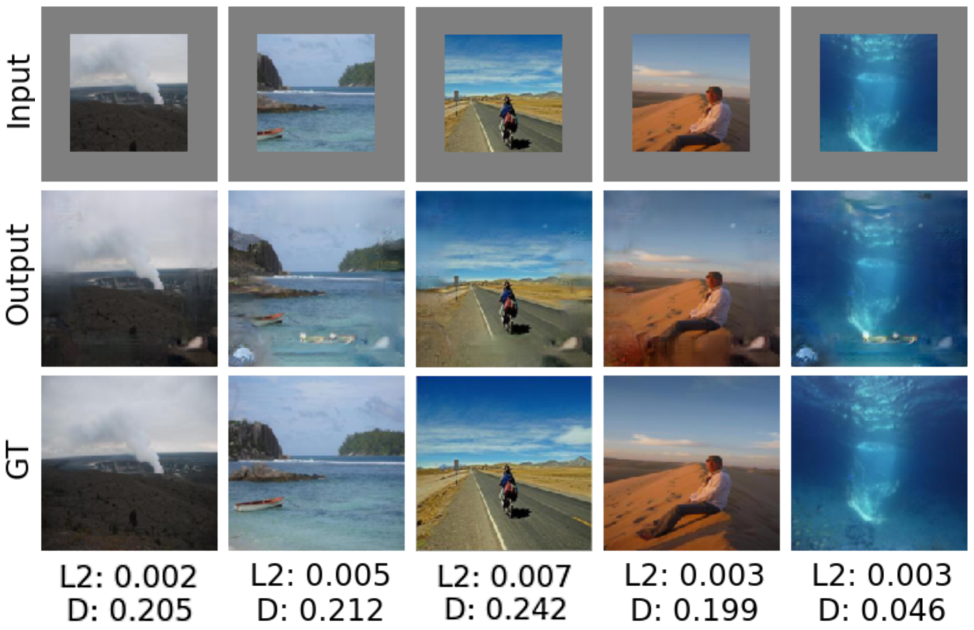}
      \caption{Low reconstruction error.}
      \vspace{0.2cm}
  \end{subfigure}%
  \begin{subfigure}[t]{0.5\textwidth}
      \centering
      \includegraphics[width=0.95\linewidth]{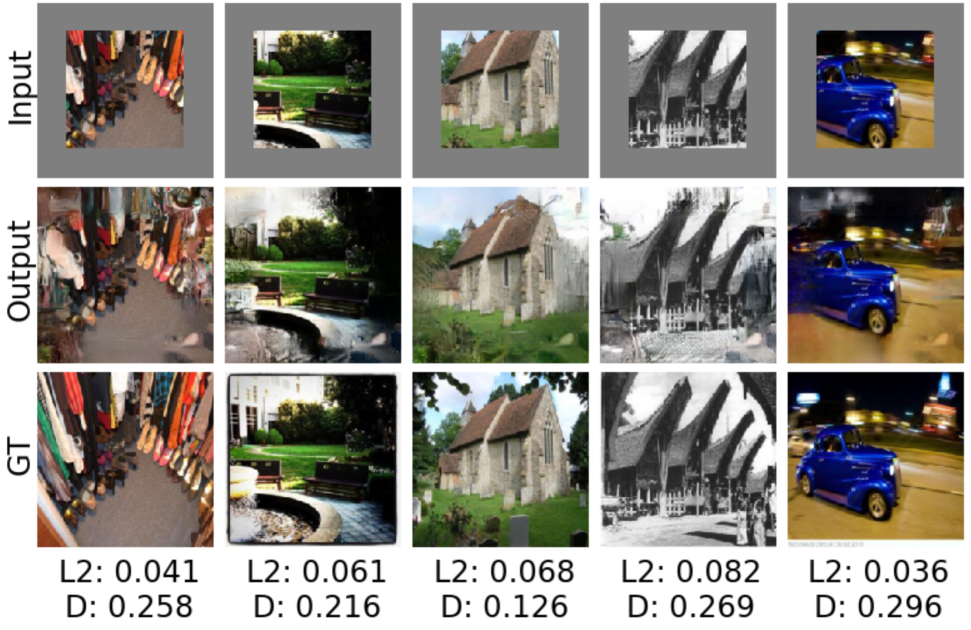}
      \caption{High reconstruction error.}
  \end{subfigure}
  \caption{Examples of outpainting on Places365 selected for varying MSEs (lower is better), revealing semantic differences within the images.}
  \label{fig:low_high_mse}
\end{figure*}

\begin{figure*}[tp]
\centering
  \begin{adjustbox}{center}
  \includegraphics[width=\textwidth]{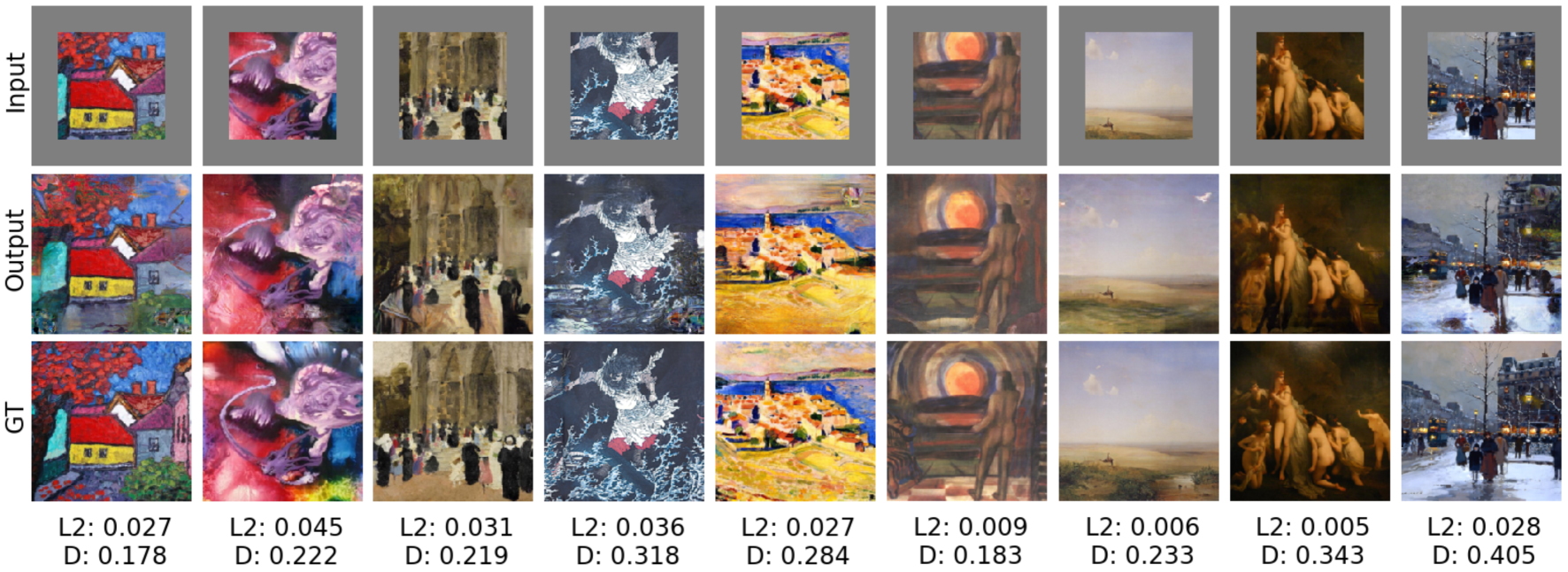}
  \end{adjustbox}
  \caption{Qualitative illustration of artwork outpainting on WikiArt.}
  \label{fig:B4_samples}
\end{figure*}

\begin{figure}[tp]
\centering
  \begin{adjustbox}{center}
  \includegraphics[width=\columnwidth]{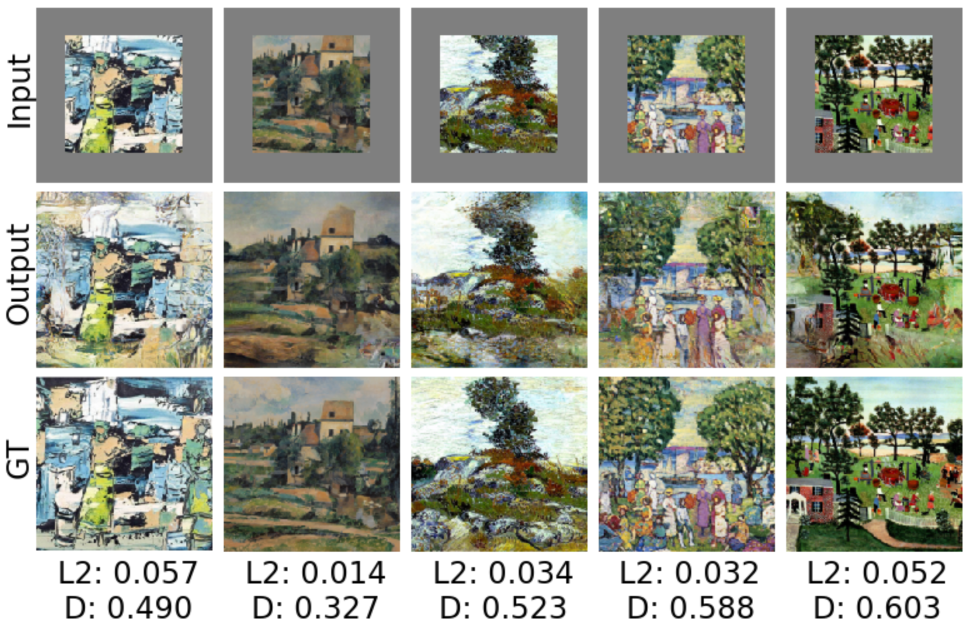}
  \end{adjustbox}
  \caption{Examples of outpainting on WikiArt selected for high discriminator output values, indicating a bias towards highly decorated paintings.}
  \label{fig:B4_high_D}
\end{figure}

\textbf{Variation in reconstruction quality} \hspace{0.3cm}
See Figure \ref{fig:low_high_mse} for samples where the MSE between the output and ground truth is either particularly low or high, when including the adversarial loss. It seems that 'smooth' images such as landscapes perform best, while highly detailed images including indoor scenes perform worst. This is understandable: if the dimensions of objects within the image are smaller than what is being omitted, they become impossible for the outpainter to predict.

\textbf{Artwork outpainting} \hspace{0.3cm}
See Figure \ref{fig:B4_samples} for a few examples of generative outpainting performed on the WikiArt validation- and test sets. In this application, we answer the question \textit{'What would the artist have drawn if he/she had a bigger canvas?'}. The results look comparatively accurate and visually pleasing, for which we think the reason is that 'imperfections' in art (to the extent that term is even well-defined) are much less bothersome than in natural imagery. It is also interesting to observe that when the output's realism $D(G(x))$ is maximized, a bias towards specific types of artwork content and/or genres can be perceived as in Figure \ref{fig:B4_high_D}.

\begin{figure*}[tp]
\centering
  \begin{adjustbox}{center}
  \includegraphics[width=0.9\textwidth]{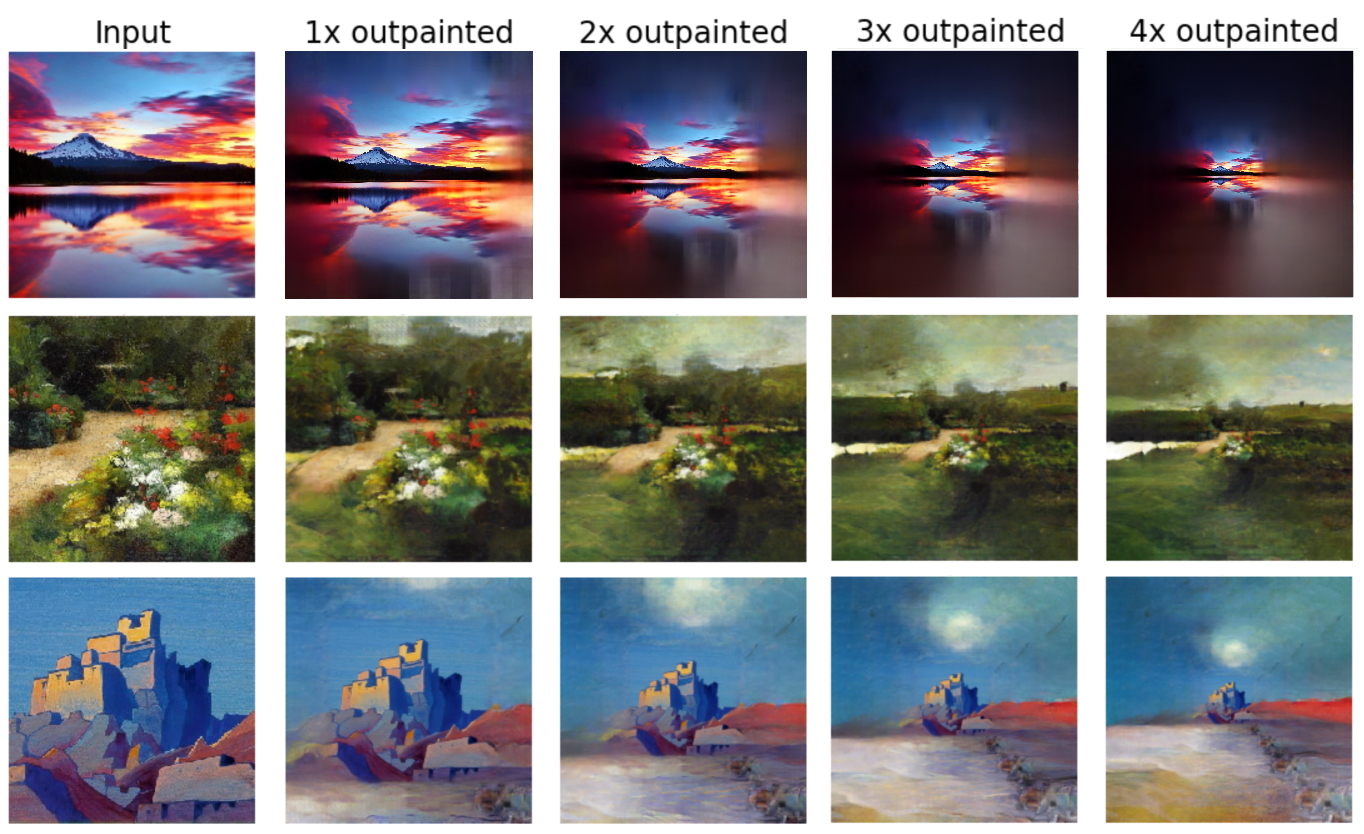}
  \end{adjustbox}
  \caption{Recursive outpainting on both natural photos and artwork, revealing some glitches but also intriguing 'eigenmodes' of the network.}
  \label{fig:B34_recursive}
\end{figure*}

\subsection{Recursive outpainting}

\hspace{\parindent} In contrast to inpainting, outpainting does not have to be performed just once; in theory, there is no limit as to how many times we can extrapolate a single image. Figure \ref{fig:B34_recursive} reveals an interesting aspect of the outpainting network: it seems that given enough iterations, the images eventually start to converge to some kind of 'eigenmode' of the generator. We moreover observe that the painting of a park with flowers seems to slowly morph into a landscape with clouds as we keep hallucinating outwards, suggesting that this scenery must appear (disproportionately) often in the training set.

\begin{figure*}[tp]
\centering
  \begin{adjustbox}{center}
  \includegraphics[width=0.8\textwidth]{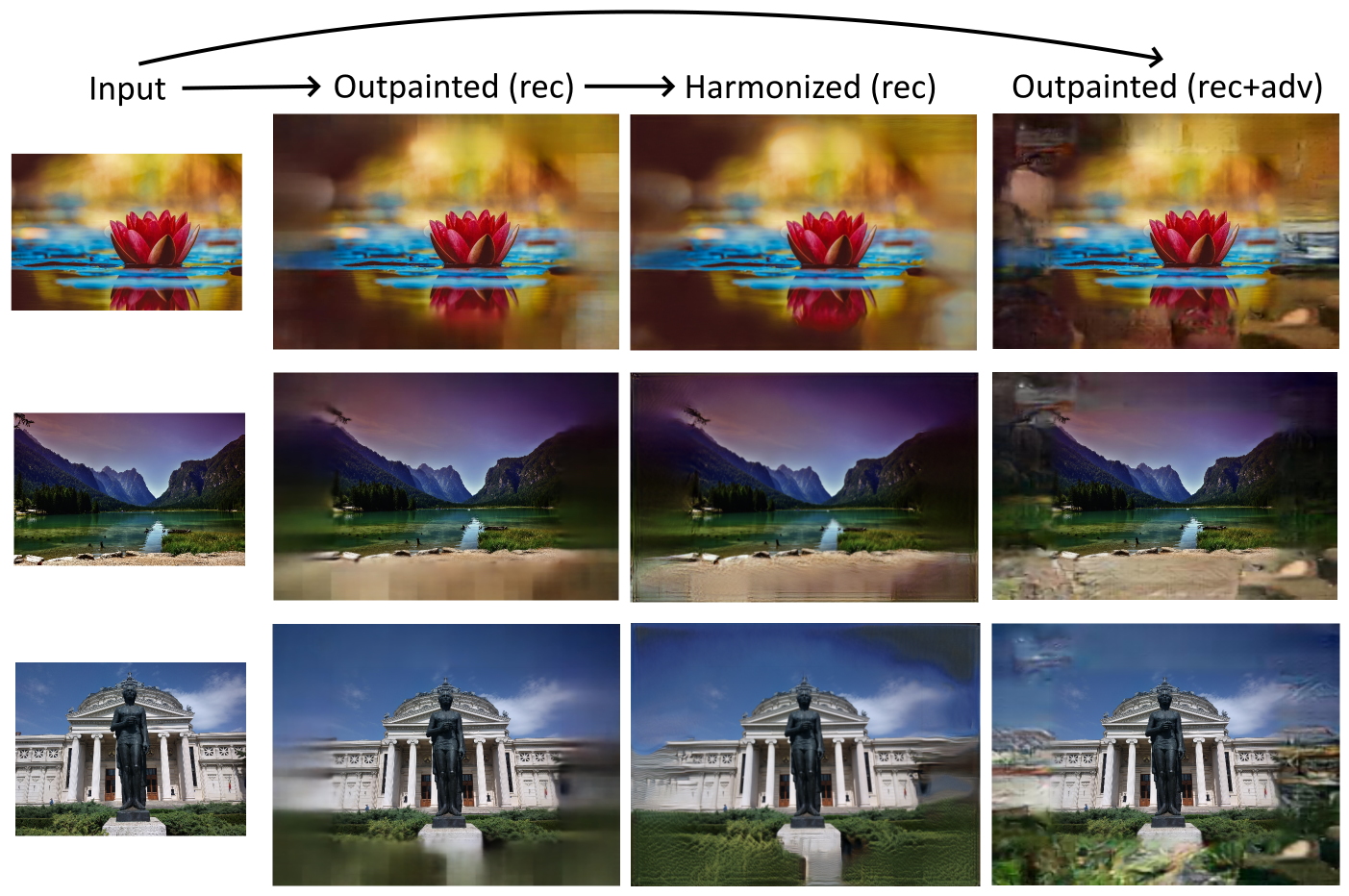}
  \end{adjustbox}
  \caption{Qualitative comparison of outpainting without adversarial loss followed by SinGAN harmonization (which has an adversarial aspect by nature), versus direct adversarial outpainting with the proposed GAN.}
  \label{fig:singan}
\end{figure*}

\subsection{Harmonization}

\hspace{\parindent} Our second method for outpainting introduces an additional step that involves employing SinGAN as described earlier. The biggest drawback of this approach is that it takes around an hour to train SinGAN on a single input image, in contrast to the outpainter itself which is a model that can simply be applied anytime once it is trained on a dataset. Nevertheless, we tested out the joint harmonization and super-resolution process on several images that demonstrate its potential, as seen in Figure \ref{fig:singan}. When applied on the output of the reconstruction loss minimizer, it has a sharpening effect. Furthermore, despite the fact that the end result still lacks photorealism, we observe fewer glitches than the first approach which uses an adversarially trained GAN directly. Arbitrarily large resolutions could in principle be supported by SinGAN, although in practice a huge amount of GPU memory ($>12$GB) is needed to scale beyond a maximum image dimension of around 500 pixels.

\section{CONCLUSION}

\hspace{\parindent} Image outpainting is a novel but exciting idea that holds promise, especially when cascaded with SinGAN to further increase the output fidelity. The models trained in this project still contain some glitches, but we believe these could be alleviated by closer investigations into the subject. Non-photorealistic images such as artwork seem to produce convincing results, which we largely attribute to human judgement becoming more permissive rather than to a better-performing model.

\section{FUTURE WORK}

\hspace{\parindent}  Other than architectural improvements, possible directions for future research include:
\begin{itemize}
  \item While enabling the discriminator in this project significantly sharpened the outpainting results, the training stability could be better and the model was not very successful in enforcing natural object semantics across the image. Instead of a single discriminator, utilizing both a global and local variant might increase the overall consistency of generated images as well as improve the realism of smaller-scale structures \cite{Iizuka2017}.
  \item At the present time, it is unknown how the model reacts to slight variations in the input space. Hence, video outpainting presents a set of new interesting challenges, mostly related to temporal consistency of the hallucinations.
\end{itemize}

\section*{ACKNOWLEDGEMENTS}

\hspace{\parindent} I would like to thank Prof. Peter Belhumeur for providing the exciting opportunity to conduct this research project, and for offering a great learning opportunity in his excellent course \textit{Deep Learning for Computer Vision}. I am also grateful to John Daciuk for giving me access to his WikiArt dataset, thus saving me time otherwise spent scraping while also allowing for aesthetically appealing experiments to be conducted more easily.

This work was performed while being the recipient of a Belgian American Educational Foundation Fellowship.


\bibliographystyle{unsrt}
\bibliography{bib}

\begin{thebibliography}{10}

\bibitem{Larsson2017}
Gustav Larsson, Michael Maire, and Gregory Shakhnarovich.
\newblock {Colorization as a proxy task for visual understanding}.
\newblock {\em Proceedings - 30th IEEE Conference on Computer Vision and
  Pattern Recognition, CVPR 2017}, 2017-Janua:840--849, 2017.

\bibitem{Doersch2015}
Carl Doersch, Abhinav Gupta, and Alexei~A. Efros.
\newblock {Unsupervised visual representation learning by context prediction}.
\newblock {\em Proceedings of the IEEE International Conference on Computer
  Vision}, 2015 Inter:1422--1430, 2015.

\bibitem{Noroozi2016}
Mehdi Noroozi and Paolo Favaro.
\newblock {Unsupervised learning of visual representations by solving jigsaw
  puzzles}.
\newblock {\em Lecture Notes in Computer Science (including subseries Lecture
  Notes in Artificial Intelligence and Lecture Notes in Bioinformatics)}, 9910
  LNCS:69--84, 2016.

\bibitem{Goodfellow2014}
Ian~J. Goodfellow, Jean Pouget-Abadie, Mehdi Mirza, Bing Xu, David
  Warde-Farley, Sherjil Ozair, Aaron Courville, and Yoshua Bengio.
\newblock {Generative Adversarial Networks}.
\newblock pages 1--9, 2014.

\bibitem{Iizuka2017}
Satoshi Iizuka, Edgar Simo-Serra, and Hiroshi Ishikawa.
\newblock {Globally and locally consistent image completion}.
\newblock {\em ACM Transactions on Graphics}, 36(4), 2017.

\bibitem{Mescheder2018}
Lars Mescheder, Andreas Geiger, and Sebastian Nowozin.
\newblock {Which training methods for GANs do actually converge?}
\newblock {\em 35th International Conference on Machine Learning, ICML 2018},
  8:5589--5626, 2018.

\bibitem{Pathak2016}
Deepak Pathak, Philipp Krahenbuhl, Jeff Donahue, Trevor Darrell, and Alexei~A.
  Efros.
\newblock {Context Encoders: Feature Learning by Inpainting}.
\newblock {\em Proceedings of the IEEE Computer Society Conference on Computer
  Vision and Pattern Recognition}, 2016-Decem:2536--2544, 2016.

\bibitem{Zhang2013}
Yinda Zhang, Jianxiong Xiao, James Hays, and Ping Tan.
\newblock {Framebreak: Dramatic image extrapolation by guided shift-maps}.
\newblock {\em Proceedings of the IEEE Computer Society Conference on Computer
  Vision and Pattern Recognition}, pages 1171--1178, 2013.

\bibitem{Sabini2018}
Mark Sabini and Gili Rusak.
\newblock {Painting Outside the Box: Image Outpainting with GANs}.
\newblock 2018.

\bibitem{Wang2019}
Yi~Wang, Xin Tao, Xiaoyong Shen, and Jiaya Jia.
\newblock {Wide-context semantic image extrapolation}.
\newblock {\em Proceedings of the IEEE Computer Society Conference on Computer
  Vision and Pattern Recognition}, 2019-June:1399--1408, 2019.

\bibitem{snapseed}
Snapseed - apps on google play, Jun 2018.

\bibitem{Shaham2019}
Tamar~Rott Shaham, Tali Dekel, and Tomer Michaeli.
\newblock {SinGAN: Learning a Generative Model from a Single Natural Image}.
\newblock may 2019.

\bibitem{Zhou2018}
Bolei Zhou, Agata Lapedriza, Aditya Khosla, Aude Oliva, and Antonio Torralba.
\newblock {Places: A 10 Million Image Database for Scene Recognition}.
\newblock {\em IEEE Transactions on Pattern Analysis and Machine Intelligence},
  40(6):1452--1464, 2018.

\bibitem{wikiart}
Wikiart: Visual art encyclopedia.

\bibitem{Huang2018}
He~Huang, Philip~S. Yu, and Changhu Wang.
\newblock {An Introduction to Image Synthesis with Generative Adversarial
  Nets}.
\newblock pages 1--17, 2018.

\bibitem{Li2018}
Haofeng Li, Guanbin Li, Liang Lin, Hongchuan Yu, and Yizhou Yu.
\newblock {Context-Aware Semantic Inpainting}.
\newblock {\em IEEE Transactions on Cybernetics}, 14(8):1--12, 2018.

\end{thebibliography}

\end{document}